# Deep Convolutional Neural Networks on Multiclass Classification of Three-Dimensional Brain Images for Parkinson's Disease Stage Prediction

Guan-Hua Huang[1,*], Wan-Chen Lai[1], Tai-Been Chen[2,3,4], Chien-Chin Hsu[5], Huei-Yung Chen[6], Yi-Chen Wu[2,6], Li-Ren Yeh[7]

[1]Institute of Statistics, National Yang Ming Chiao Tung University, Hsinchu, Taiwan

[2]Department of Medical Imaging and Radiological Sciences, I-Shou University, Kaohsiung, Taiwan

[3]Department of Radiological Technology, Faculty of Medical Technology, Teikyo University, Tokyo, Japan

[4]Infinity Co. Ltd., Taoyuan, Taiwan

[5]Department of Nuclear Medicine, Kaohsiung Chang Gung Memorial Hospital, Kaohsiung, Taiwan

[6]Department of Nuclear Medicine, E-Da Hospital, I-Shou University, Kaohsiung, Taiwan

[7]Department of Anesthesiology, E-Da Cancer Hospital, I-Shou University, Kaohsiung, Taiwan

[*]Corresponding author: Guan-Hua Huang, Institute of Statistics, National Yang Ming Chiao Tung University, No. 1001, Daxue Rd., East Dist., Hsinchu City, 300093, Taiwan
E-mail: ghuang@nycu.edu.tw

**Abstract**

Parkinson's disease (PD), a degenerative disorder of the central nervous system, is commonly diagnosed using functional medical imaging techniques such as single-photon emission computed tomography (SPECT). In this study, we utilized two SPECT data sets (n = 634 and n = 202) from different hospitals to develop a model capable of accurately predicting PD stages, a multiclass classification task. We used the entire three-dimensional (3D) brain images as input and experimented with various model architectures. Initially, we treated the 3D images as sequences of two-dimensional (2D) slices and fed them sequentially into 2D convolutional neural network (CNN) models pretrained on ImageNet, averaging the outputs to obtain the final predicted stage. We also applied 3D CNN models pretrained on Kinetics-400. Additionally, we incorporated an attention mechanism to account for the varying importance of different slices in the prediction process. To further enhance model efficacy and robustness, we simultaneously trained the two data sets using weight sharing, a technique known as cotraining. Our results demonstrated that 2D models pretrained on ImageNet outperformed 3D models pretrained on Kinetics-400, and models utilizing the attention mechanism outperformed both 2D and 3D models. The cotraining technique proved effective in improving model performance when the cotraining data sets were sufficiently large.

# Introduction

Parkinson's disease (PD) is a neurodegenerative disease with a typical onset age of 65 to 70 years; PD is characterized by its main motor symptoms, namely bradykinesia, rigidity, and tremor, but the disease also has additional motor and nonmotor characteristics [1]. PD occurs when nerve cells, or neurons, in a brain area that controls movement become impaired or die. In 2011, the incidence rate of PD per 100,000 in Taiwan was 85.1 and 364.1 for those aged 60–69 and ≥80 years, respectively [2].

According to the level of clinical disabilities or symptoms, PD can be divided into several stages [3, 4]. However, individuals with PD may exhibit differences in the type and severity of symptoms despite being at the same stage. Anatomic–pathologic studies have highlighted the high misdiagnosis rate for PD [5]; thus, functional imaging techniques are increasingly being used to understand the pathophysiology and evolution of PD. Common functional imaging techniques used for this purpose include positron emission tomography, single-photon emission computed tomography (SPECT), and functional magnetic resonance imaging. The images we used in this study were obtained using SPECT.

In clinical practice, SPECT images are typically evaluated visually or through region-of-interest (ROI) analysis. The ROI approach involves marking out or positioning ROIs over the striatum (target) and the occipital cortex (reference) and subsequently applying the background subtracted striatal uptake ratio as the evaluation metric [6]. Another approach involves conducting shape and intensity distribution analysis, with pattern recognition techniques being used for differentiation [7, 8].

Computer-aided diagnosis of PD has been researched from various perspectives,

but most studies have focused on the pattern recognition approach. Generally, the overall process involves two parts: feature extraction and classification. Feature extraction can be achieved (1) by applying dimension reduction approaches such as principal component analysis or singular value decomposition on the voxels of the whole brain [9, 10], or (2) by using the voxels of the striatum or the striatal binding ratio values as features [11-14]. After feature extraction, machine learning classifiers including support vector machine, linear/quadratic discriminant analysis, or naïve Bayes can be applied [9, 10, 12, 14-16].

In traditional machine learning classifiers, features are typically extracted using handcrafted rules. In contrast, deep learning methods such as convolutional neural networks (CNNs) can automatically extract sufficient and representative features from images, enabling automation in the computer-aided diagnosis process. CNNs are increasingly being adopted for predicting Parkinson's Disease (PD) [17-19]. While several approaches have been developed for PD diagnosis, most focus on binary classification—merely indicating whether an image shows signs of PD. Although PD is typically classified into several stages [3], multiclass classification approaches are rare [20]. Additionally, many studies on PD classification have used two-dimensional (2D) images despite the availability of three-dimensional (3D) information in SPECT data [20]. When 2D models are applied to 3D SPECT data, 2D image slices with the clearest striatum shape are selected for analysis. However, this selection process must be performed by physicians or experts, and it fails to capture the temporal or positional information that 3D SPECT images provide. To reduce the labor involved in manual selection and to utilize all available 3D information, models that take whole-brain images as inputs are urgently needed. Therefore, we developed a fully automated process that uses the entire 3D brain image as input for the multiclass classification of PD stages.

Three-dimensional images can be viewed as a series of 2D image slices, allowing them to be analyzed by 2D CNN models. In this approach, all slices pass through shared 2D convolutional layers to generate image representations, which are then combined for final PD stage prediction [21, 22]. Given that we are working with 3D medical images, using 3D CNN models is also appropriate to fully leverage the available information [23, 24]. There has been considerable debate about the effectiveness of 2D versus 3D representation learning for 3D medical images. While 2D methods benefit from extensive pretraining, they cannot fully capture the 3D context. Conversely, 3D methods excel at understanding 3D contexts but lack large, diverse datasets for effective pretraining [25]. This paper introduces an architecture that considers the relationships among 2D slices, enabling models to learn the different contributions of each slice. Additionally, we co-train two SPECT datasets from different hospitals using weight sharing to improve the model's effectiveness and robustness. We include demographic features (age and sex) to enhance prediction accuracy, particularly in multiclass classification, which is not commonly explored in similar works.

This paper provides an extensive experimental evaluation of multiple approaches for 3D medical image classification. By systematically comparing these approaches, our study offers a holistic understanding of the trade-offs between 2D and 3D models, the importance of inter-slice relationships, and the utility of cotraining and demographic features. This extensive experimental framework not only strengthens the rigor of our findings but also provides a roadmap for future research in 3D medical image classification.

## Materials

We used two data sets provided by two different hospitals: (1) the Kaohsiung Chang Gung Memorial Hospital (hereinafter called Chang Gung) and (2) the E-Da Hospital (hereinafter called E-Da). For the Chang Gung data set, a total of 634 patients who underwent brain SPECT/CT scans in the Department of Nuclear Medicine at Kaohsiung Chang Gung Memorial Hospital, Taiwan, between January 2017 and May 2018, were included in this retrospective study. The study was approved by the Chang Gung Medical Foundation Institutional Review Board, with a waiver for written informed consent due to its retrospective nature. For the E-Da data set, SPECT images and diagnostic reports from patients scanned between March 2006 and August 2013 were retrieved from the picture archiving and communication system at E-Da Hospital, I-Shou University, Taiwan. After excluding ineligible cases, 202 patients were included in the final analysis. This study was approved by the Institutional Review Board of E-Da Hospital, and all participants provided written informed consent. All patient images and associated data from both data sets were fully deidentified before inclusion in the study, in compliance with applicable regulations and institutional policies.

Two data sets saved in the DICOM (digital imaging and communications in medicine) files contained data on SPECT imaging using $^{99m}$Tc-TRODAT-1 with three dimensions $(T, H, W)$, where $T$ represents the number of slices for each person, and $H$ and $W$ represent the height and width of each slice, respectively. The three views (axial, coronal, and sagittal) of our data are presented in Fig. 1. In both data sets, each slice (axial view) was in the form of $(H, W) = (128, 128)$, with $T$ varying from person to person. The bar plots of the numbers of slices for the two data sets are illustrated in Fig. 2. The image quality differed between these two data sets. We randomly selected one original image slice, not subjected to augmentation, from each data set (Fig. 3); compared with the image from the E-Da data set, the one from the Chang Gung data set had a higher resolution and the proportion of the image occupied by the brain was also larger.

[Fig. 1 about here.]

[Fig. 2 about here.]

[Fig. 3 about here.]

The two studies were conducted independently, resulting in differences in how PD illness stages were characterized: the Chang Gung data set used four illness stages including healthy cases and the E-Da data set used six stages.

The Chang Gung data set contained 634 samples, with PD illness stages being evaluated by three experienced board-certified nuclear medicine physicians. Blinded to patient clinical symptoms and histories, the disease labels we used were determined by the majority vote of the three physicians' evaluation based on the SPECT images, yielding 174 normal samples and 460 samples with PD. The illness of those with PD was further classified into three stages with the number of patients in stages I (nearly normal), II (potentially abnormal), and III (abnormal) being 127, 77, and 256, respectively [26]. Each SPECT image also included information on age, sex, and the indices of nine selected slices that physicians used for diagnosis. Other information about the image and the tracer was recorded in the DICOM file.

The E-Da dataset consisted of 6 healthy individuals and 196 patients with PD confirmed by the physicians from the E-Da hospital. According to patient diagnostic reports, the physical symptoms of PD were classified into five different stages using the Hoehn and Yahr scale [3]. The numbers of samples for stages I (mild) to V (most severe) were 22, 27, 53, 87, and 7, respectively [20]. We anticipated having problems related to imbalanced data during subsequent data analysis. Furthermore, the indices of the selected slices that contained the clearest striatum shape were provided. Other information such as age and sex was recorded in the DICOM file.

## Methods

### Preprocessing

**Normalization**

Because the ranges of the pixel intensity values for each patient differed, potentially affecting the efficiency of machine learning algorithms, we scaled them by using min–max normalization *individually*:

$$X_i(\text{norm}) = \frac{X_i - \min(X_i)}{\max(X_i) - \min(X_i)}$$

where $X_i$ represents the pixel values in all slices of the $i$th patient, and $\min(X_i)$ and $\max(X_i)$ are the smallest and largest pixel value of $X_i$, respectively. After rescaling was conducted, the values all ranged from 0 to 1, and the relative magnitude among patients could also be maintained.

**Slice Selection**

For the Chang Gung data set, during diagnosis, physicians had selected the slices whose striatum can be recognized most clearly. The data set contained the indices provided by physicians for selected slices, with nine slices being selected for each patient. As for the E-Da data set, because physicians did not provide the information about which slices they considered, we used the slices selected by the authors from our previous work [20], where one slice that contained the clearest striatum shape was selected for each patient.

We developed an automatic process to filter all *incomplete* slices; that is, slices where parts of the image were masked, or the pixel values were all very close to zero. These slices provided no or little information, or they may disturb the training process. Although the slice size in both data sets was $128 \times 128$ pixels, the proportions that the brain accounted for were different: the proportions in the Chang Gung data set were larger than those in the E-Da data set (Fig. 3). We thus removed the slices whose number of pixels with normalized intensity value greater than 0.1 was less than 800 and 400 in the Chang Gung and E-Da data sets, respectively.

**Other Information**

Epidemiological studies have indicated the existence of sex differences in PD: the incidence and prevalence of PD in men are higher than those in women [27]. PD is also related to age: the incidence rates rise rapidly after the age of 60 years [28]. Moreover, the male:female incidence rate ratio increases with age in general. For the aforementioned reasons, age and sex were considered during the training process.

Because ages in this data set were positive numbers in years, they needed to be normalized to range $[0, 1]$ to ensure that the scale would be the same as that of the pixel values of slices. The normalized age is expressed as

$$a'_i = a_i/100, \quad a_i \in [0, 100],$$

where $a_i$ represents the age in years of the $i$th patient.

Sex only took two values, male or female; thus, it was represented by a dummy variable defined as

$$G = \begin{cases} 1 & \text{if gender is male} \\ 0 & \text{if gender is female} \end{cases}.$$

**Augmentation**

Because the two data sets were both imbalanced and too small for deep learning models, overfitting was a concern; that is, the model may fit the training data excessively, consequently leading to poor performance when predicting new data. To avoid this problem, the most common approach for image classification tasks is to perform data augmentation, which provides new data by making minor alterations to the existing dataset. The augmented data were generated using video transform and depth transform, as described in the following section. We used online augmentation; that is, image augmentation was conducted in mini-batches before feeding data to the model. The model with online augmentation was presented with different images at each epoch; this aided the generalization ability of the model, and because the model did not need to save the augmented images on the disk, the computational burden was reduced.

**Video Transform**

Because our data were 3D SPECT images, which can be regarded as a series of slices (2D images), we adopted video transforms to ensure that the same random parameters (e.g., crop size, rotation angle, flip or not) would be applied to all the slices of each patient:

- Random rotation (degree = 5): rotate the image by an angle ranging from $-5°$ to $5°$
- Center crop (size = (72, 72)): crop the given image at the center, with the desired output size being $72 \times 72$
- Resize (size = (72, 72)): resize images to $72 \times 72$

**Depth Transform**

Video transform yielded images with a size of $72 \times 72$; however, the number of slices for each patient was different. We applied the trilinear interpolation approach [29] to construct new slices to unify the depth dimension of all patients' images. Trilinear interpolation is a multivariate interpolation method applied to 3D inputs (i.e., volumetric data sets). The value is approximated linearly by the intermediate point within the nearest cubic lattice, which is identical to two bilinear interpolations combined with a linear interpolation.

Notably, the result of trilinear interpolation is unrelated to the order of the steps applied along the three axes. We set the target slice number to 32 because most of the samples in our data sets had fewer than 32 slices, and thus we would not lose too much information. After we applied trilinear interpolation, the data size of each patient was unified as $32 \times 72 \times 72$.

**Multiclass Classification**

The main objective of this study was to develop a valid model to yield accurate prediction of PD illness stages. Because each sample can only belong to one of $C$ PD stages (including healthy) ($C = 4$ for the Chang Gung data set and $C = 6$ for the E-Da data set), this was a multiclass classification task. We conducted a deep learning model that mapped inputs of the $i$th image $\mathbf{X}_i$ to a $C$-dimentional label vector $\mathbf{y}_i = (y_{i1}, y_{i2}, \cdots, y_{iC})$ with $\forall y_{ic} \in \{0,1\}, y_{i1} + \cdots + y_{iC} = 1$. Therefore, the categorical cross-entropy can be used as the loss function:

$$\mathcal{L}_{\text{CE}} = \sum_i \left\{ \sum_{c=1}^{C} \left[ -y_{ic} \ln\left(s(f_c(\mathbf{X}_i))\right) \right] \right\},$$

where $f_c(\boldsymbol{X}_i)$ is $\boldsymbol{X}_i$'s $c$th input for the final fully connected layer and $s(z_c) = e^{z_c}/\sum_{j=1}^{4} e^{z_j}$ is the softmax function.

Our datasets were highly imbalanced. Most machine learning algorithms for classification problems operate under the assumption of an equal number of samples in each class. In an unbalanced data set, the learning can be biased toward the majority classes and fail to catch the patterns of the minority classes. A popular method to address imbalanced data is to set the *class weight* in the loss function. We defined our class weights as

$$w_c = \frac{N_c}{\sum_{j=1}^{6} N_j} \text{ with } N_c = \frac{n}{n_c}, c = 1, \cdots, C,$$

where $n$ represents the total number of samples, and $n_c$ is the number of samples in the $c$th class (PD stage). The loss function with class weights thus becomes

$$\mathcal{L}_{\text{WCE}} = \sum_i \left\{ \sum_{c=1}^{C} \left[ -w_c y_{ic} \ln\left(s(f_c(\boldsymbol{X}_i))\right) \right] \right\}.$$

**Deep CNN Models**

Image classification using CNN has demonstrated outstanding performances compared with traditional machine learning approaches. CNNs are designed to automatically and adaptively learn spatial hierarchies of features through back-propagation over multiple stacked building blocks, such as convolution layers, non-linear layers (activations), pooling layers, and fully connected layers. CNN is a deep learning method designed to process structured arrays, and it has become dominant in various computer vision tasks. In addition to their application to classification, segmentation, and recognition problems related to image or video, CNNs have also been applied to natural language processing. CNNs are designed to automatically and adaptively learn spatial hierarchies of features through back-propagation over multiple stacked building blocks, such as convolution layers, non-linear layers (activations), pooling layers, and fully connected layers. Generally, CNNs are composed of multiple layers of artificial neurons that are mathematical functions for calculating the weighted sum of multiple inputs and then yielding an activation value. Each neuron behaves according to its weights. The operation of multiplying pixel values by weights and summing them is called *convolution*. When a CNN is fed the pixel values of an image, each layer generates several activation maps that highlight the relevant features of the input image. The early few (or bottom) layers usually detect more basic features such as edges and corners, with deeper layers extracting higher-level features such as objects. Lastly, based on the output of the final convolution layer, the fully connected layers output a set of scores ranging from 0 to 1 that indicate how likely it is that the input belongs to each class.

This study focuses on multiclass classification of 3D SPECT images. Traditionally, 3D images are analyzed using either 2D CNNs on individual slices (2D models) or 3D CNNs on the entire spatiotemporal volume (3D models). We introduce an architecture that uses an attention mechanism to consider the relationships among slices, allowing each slice to contribute differently to the prediction (slice-relation-based models). Besides training the Chang Gung and E-Da data sets separately with various model architectures, we propose cotraining the two datasets simultaneously using weight sharing to enhance model effectiveness and robustness.

Transfer learning is the process of creating new models by fine-tuning previously trained networks, where a model trained for one task is utilized as the initial value of the model for another related task. That is, the knowledge gained from previous tasks is transferred to related ones. Transfer learning has been widely applied to diverse tasks. One of its strengths is particularly helpful for our task: transfer learning reduces the need and effort to recollect a large amount of training data, thus addressing the challenge of training a full-scale model from scratch or with little data. In our transfer learning training process, we used 2D and 3D model architectures trained on ImageNet [30] and Kinetics-400 [31], respectively, as the pretrained model, froze the weights of the first few layers, set the remaining layers as trainable, and then replaced the final few layers with customized layers. Given that ImageNet and Kinetics-400 were trained on natural RGB images, some studies demonstrated that, in different medical imaging applications, CNNs pretrained on these large-scale natural image data sets performed better and were more robust to the size of the target training data than the CNNs trained from scratch [32, 33].

**2D Models**

3D SPECT data can be regarded as a series of 2D images and thus can be analyzed by 2D model architectures. We selected the VGG-16 architecture pretrained on ImageNet as our pretrained model, and we used its convolution layers connected with some customized layers to form our training model.

Because the images in ImageNet were RGB images with three channels, we copied each slice three times to generate the three-channel input from our gray-scale medical images, where each input was formed by *a set of three repeated slices*. All slices passed through shared VGG-16 convolutional layers to obtain image representations; these outputs were then "summarized" through some customized layers before proceeding to the final fully connected (FC) layers. We discarded the last max-pooling layer in the convolution layers of the original VGG-16 architecture to ensure that the output shape would not be too small. The different customized layers we used are described in the following sections.

*VGG plus linear (Linear)*: The first model was the simplest one and also the baseline model we used for comparison. We added an adaptive average pooling layer (with target output size $1 \times 1$) to features with size $(512, 4, 4)$ and obtained outputs of size $(512, 1, 1)$. After the outputs from all slices were averaged, a multilayer perceptron (MLP) with rectified linear unit (ReLU) activation was used to reduce the output dimension from 512 to 16. Age and sex were concatenated with the output features if necessary and then fed into the FC layers. In this manner, the effects of age and sex would be appropriately considered. The model architecture is presented in Fig. 4 (a).

[Fig. 4 about here.]

*VGG plus Conv2D (Conv2D)*: For more advanced processing, we replaced the average pooling layer in the aforementioned *Linear* model architecture with a 2D convolutional layer. This enabled us to extract more realistic features. This model architecture is presented in Fig. 4 (b).

*Axial–coronal–sagittal convolutions (ACS)*: Because any of the three views (axial, coronal, and sagittal) of our 3D data can be regarded as 2D images, ACS convolutions [25] were utilized in this model. In ACS convolutions, 2D kernels are split by the channel into three parts and convoluted separately on the axial, coronal, and sagittal views of 3D inputs; as a result, the weights pretrained on large 2D data sets could still be used. Suppose we wished to obtain a 3D output $X_o \in \mathbb{R}^{C_o \times T_o \times H_o \times W_o}$ with pretrained 2D kernels $W \in \mathbb{R}^{C_o \times C_i \times K \times K}$, given a 3D input $X_i \in \mathbb{R}^{C_i \times T_i \times H_i \times W_i}$, where $C_i$ and $C_o$ were the input and output channels, respectively, $T_i \times H_i \times W_i$ and $T_o \times H_o \times W_o$ denoted the number of slices, height, and width of the input and output shapes, respectively, and $K$ represented the kernel size. To transfer the 2D kernels to 3D kernels, we *unsqueezed* the 2D kernels into pseudo 3D kernels on an axis [34]. The ACS convolutions functioned by splitting and unsqueezing the 2D kernels into three parts that aimed for representations of three views: $\boldsymbol{W}_a \in \mathbb{R}^{C_o^{(a)} \times C_i \times K \times K \times 1}$, $\boldsymbol{W}_c \in \mathbb{R}^{C_o^{(c)} \times C_i \times K \times 1 \times K}$, and $\boldsymbol{W}_s \in \mathbb{R}^{C_o^{(s)} \times C_i \times 1 \times K \times K}$, where $C_o^{(a)} + C_o^{(c)} + C_o^{(s)} = C_o$. The output feature of each view then became

$$\boldsymbol{X}_o^{(v)} = \text{Conv3D}(\boldsymbol{X}_i, \boldsymbol{W}_v) \in \mathbb{R}^{C_o^{(v)} \times T_o \times H_o \times W_o}, v \in V = \{a, c, s\}.$$

$\boldsymbol{X}_o^{(a)}$, $\boldsymbol{X}_o^{(c)}$, and $\boldsymbol{X}_o^{(s)}$ were concatenated after an average pooling layer to obtain the overall output feature, and finally went through the FC layers (Fig. 4 (c)).

**3D Models**

Because we were working with 3D medical images, the application of 3D models was also appropriate. Here we employed the 3D models for action recognition and video classification, as in Tran et al. [23], based on the ResNet-18 architecture. These models were all pretrained on the Kinetics-400 dataset [31], containing 306245 short-trimmed, realistic action clips from 400 action categories.

Notably, the input shape for these 3D models becomes $(C, T, H, W)$, where $C$ is the channel and $T$ is the number of video frames in a clip. All of our slices were gray-scale with one channel, whereas each frame of a clip in Kinetics-400 was an RGB image with three channels. For data set compatibility, each of our slices was repeated three times, and then video transform and depth transform were applied to ensure that all the inputs had the same shape.

The overall architectures of the following 3D models were similar: a 3D pretrained model was followed by an average pooling layer and some MLPs. However, different pretrained models were used for each 3D model.

*3D ResNet (R3D)*: Unlike 2D CNNs, 3D CNNs can preserve and capture temporal, or positional in our case, information because the filters are convoluted over both time and space dimensions. In this model, the 2D convolutions were merely replaced with 3D convolutions, with an extra temporal extent on the filter. The overall model architecture is illustrated in Fig. 5 (a).

[Fig. 5 about here.]

*Mixed convolutions ResNet (MC3)*: On the basis of the hypothesis that temporal modeling with 3D convolutions would be useful only in the early layers and would be unnecessary in the late layers because the higher-level features are abstract [23], the mixed 2D–3D convolutions were used. Considering that the *R3D* model had five groups of convolutions, we applied the *MC3* model where the 3D convolutions in groups 3, 4, and 5 were replaced with 2D convolutions. The architecture is presented in Fig. 5 (b).

*(2+1)D ResNet (R(2+1)D)*: A 3D convolution can be approximated by a spatial 2D convolution followed by a temporal one-dimensional (1D) convolution, where the model captures spatial and temporal information from two separate steps. This process facilitates the optimization and doubles the number of nonlinear transformations caused by the additional activations between 2D and 1D convolutions without increasing the number of parameters. In this model, the (2+1)D convolutions are substituted for 3D convolutions, and the architecture is illustrated in Fig. 5 (c).

**Slice-Relation-Based Models**

We initially assumed that all the slices from one patient contributed equally in the training process of the 2D models because we merely used the average of their convoluted features. However, during diagnosis, physicians only focus on slices whose striata are clear enough to be recognized; thus, treating all the slices of one patient equally is not ideal. We therefore also considered the relation among slices to assist models to learn the differences among slice contributions.

*Index embedding (IdxEmb)*: First, the slices of one patient were sequentially numbered according to the order of their entry into the model, called *slice index*. The slice index was regarded as a categorical variable $x$ and then mapped to a vector $\boldsymbol{x}$ with a predefined dimension through *entity embedding* [35]. Entity embedding was used to operate a linear layer on the one-hot encoding of $x$ as follows:

$$\boldsymbol{x} \equiv \boldsymbol{W}\boldsymbol{\delta}_x$$

where $m$ is the number of possible values for the categorical variable $x$, $\boldsymbol{\delta}_x$ is a vector of length $m$ with the $i$th element being 1 if $i = x$ and 0 otherwise, $i = 1, \cdots, m$, $\boldsymbol{W} = \{w_{ij}\} \in \mathbb{R}^{k \times m}$ is the weight matrix that connects the one-hot encoding with the entity embedding, and $k$ is the dimension of the entity embedding. $\boldsymbol{W}$ can be regarded as the weights of the linear layer, which were trainable by using a standard backpropagation method. The embedded index $\boldsymbol{x}$ was reshaped and added to the convoluted features of the corresponding slice and went through a 2D convolution together. The output features of these slices were then gathered and averaged and finally fed into a classifier. The whole model architecture is presented in Fig. 6 (a).

[Fig. 6 about here.]

*Attention (Attn)*: Adding the information about the position of slices in the slice index embedding model seemed appropriate, but the image registration problem proved challenging: we should have the absolute position for slices. Unfortunately, our data were functional images whose positional information was inaccurate; for example, slice index 1 might correspond to different brain areas for different patients. To avoid the image registration problem, we only considered the relative importance of all slices in one patient, an approach motivated by the *attention* mechanism [36, 37]. Because our attention mechanism only considered the input images of a patient with no extra output information and aimed for capturing the internal correlation of slices within a patient, we were in fact implementing *self-attention* [38, 39].

Overall, we adopted the 2D model but, instead of taking the average, we used the weighted sum (Fig. 6 (b)). Details for the weighted sum are as follows. For one patient, each VGG-16 convolution output feature $g_i$ went through a convolution layer with ReLU activation and yielded another output $h_i$ of size $(1,1,1)$. We applied the softmax function to outputs $h_1, \cdots, h_m$ to form the weight for each output feature $g_i$: $w_i = e^{h_i} / \sum_{j=1}^{m} e^{h_j}$, where $m$ is the number of slice sets. Subsequently, the input of the following MLP (with the ReLU activation) can be represented as follows: $u = \sum_{i=1}^{m} w_i g_i$. In this manner, the model allowed each patient to have a different weighting scheme (i.e., different $w_1, \cdots, w_m$) with unique important slices.

*Multihead Attention (MH-Attn)*: The aforementioned attention mechanism focuses on a specific aspect that the image slices of a patient reflected. However, multiple aspects in these image slices together form the overall structure of the 3D image. Multihead attention allows the model to jointly attend to information from different aspect subspaces for different slices. We here adopted the multihead attention formulation used in Vaswani et al. [39].

After going through the VGG-16 convolutional layers and an average pooling layer, each output feature $f_i$ (with dimension 512) had three (randomly initialized) weight matrices to be multiplied separately: query weights $W^Q$, key weights $W^K$, and value weights $W^V$. The multiplications then yielded $q_i$ (query), $k_i$ (key), and $v_i$ (value). Thus,

$$Q = FW^Q, K = FW^K, V = FW^V,$$

where $W^Q, W^K, W^V \in \mathbb{R}^{512 \times 512}$, $F = (f_1, \cdots, f_m)^T$, $Q = (q_1, \cdots, q_m)^T$, $K = (k_1, \cdots, k_m)^T$, and $V = (v_1, \cdots, v_m)^T$. Each query $q_i$ multiplied the key matrix $K$ to generate its attention scores. These attention scores went through a softmax function to create the weights for the weighted sum over $v_1, \cdots, v_m$, where the weighted sum was the attention vector $z_i$. The aforementioned process can be expressed as

$$\text{Attention}(Q, K, V) = \text{softmax}(QK^T)V = Z,$$

where $Z = (z_1, \cdots, z_m)^T$. One set of $(W^Q, W^K, W^V)$ matrices was called an attention *head*. Multiple attention heads enable the model to learn different relevance among slices. We used four heads (based on our empirical results), and the independent outputs of these heads were simply concatenated and transformed into the desired dimension 512 by using matrix multiplication. The multihead attention mechanism can be expressed as

$$\text{MultiHead}(Q, K, V) = \text{concatenate}(\text{head}_1, \cdots, \text{head}_4)W^O,$$

where

$$\text{head}_i = \text{Attention}(FW_i^Q, FW_i^K, FW_i^V), i = 1, \cdots, 4,$$

and $W^O \in \mathbb{R}^{2048 \times 512}$. The output shape became $(m, 512)$ after the application of this multihead attention mechanism. The sum of these row vectors then served as the input to the following MLP (with ReLU activation). The overall model architecture is illustrated in Fig. 6 (c).

**Cotraining Models**

Two datasets used in this study differ significantly in how they define PD illness stages: the Chang Gung dataset includes four stages (including healthy cases), while the E-Da dataset uses six stages. This discrepancy makes it challenging to directly train on one dataset and validate on the other without additional alignment or adjustments.

We initially trained our two datasets (Chang Gung and E-Da data sets) separately with different model architectures. Although we obtained some favorable results, we still did not leverage the high similarity between them: they were both SPECT brain images and thus some shared characteristics should be present. Thus, we subsequently trained these two data sets jointly by using the models introduced in the preceding sections. The schematic is illustrated in Fig. 7. The 2D and 3D pretrained models were still used, but the inputs were evenly composed of the two data sets in each batch. The shared model weights (i.e., the yellow block in Fig. 7) were first cotrained using inputs evenly distributed from both datasets, ensuring that the model could learn features applicable across datasets. The number of cotraining layers was a hyperparameter to be tuned. After this shared training phase, the remaining components of the model were trained separately for each dataset, optimizing for their respective loss functions to account for the differences in stage definitions. The loss $\mathcal{L}$ can be expressed as

$$\mathcal{L} = \mathcal{L}_{\text{CG}} + \mathcal{L}_{\text{EDA}}$$

where $\mathcal{L}_{\text{CG}}$ and $\mathcal{L}_{\text{EDA}}$ refer to the weighted categorical cross-entropy loss of the Chang Gung and E-Da data sets, respectively. In this manner, the two data sets would share the weights of the cotraining part, which increased the robustness of the model because of the increased variation and number of training samples. The model could also retain the differentiation between the two data sets through the parts trained separately.

[Fig. 7 about here.]

## Evaluation

### K-Fold Cross-Validation

To assess the generalization ability of the models and to avoid overfitting and selection bias, we adopted stratified cross-validation. The number of folds was set to be five, which meant the original samples were randomly partitioned into five groups, with each group containing approximately the same proportions of class labels. In each cross-validation round, one group was for testing (called the test set), and the other four groups were for training (called the training set); 20% of the training set served as the validation set.

### Metrics

We used two metrics, namely accuracy and F1 score, to evaluate our model performance. Accuracy is defined as the proportion of samples that are correctly classified. This evaluation metric is commonly used in classification tasks. However, a problem with imbalanced data sets is that model accuracy can be judged as high even if it only correctly predicts the majority class. Because our data sets were imbalanced, we employed another metric to evaluate model performance: F1 score. For each class, we are interested in the fraction of relevant samples among the retrieved samples (i.e., precision) and the fraction of retrieved samples among the relevant samples (i.e., recall). The F1 score is the harmonic mean of the precision and recall, with its optimal and poorest values being 1 and 0, respectively. In our task, we first calculated the F1 score for each class and then took the average of these F1 scores, called macro F1 score, as our alternative evaluation metric.

# Results

This section contains the results of all the experiments performed in the current study. Table 1 summarizes our experimental configurations and parameter settings.

[Table 1 about here.]

## 2D Models

We first conducted experiments with a different number of selected slices for the analysis. Different 2D models were then applied to our two data sets separately and jointly.

### Number of Slices Used

The results are illustrated in Table 2. For the Chang Gung data set where the indices of the mine slices selected by the physicians for each patient were available, we adopted the fifth slice for the *Selected 1* experiment and the fourth, fifth, and sixth slices for the *Selected 3* experiment. For the E-Da data set where only one slice was selected for each patient, the three consecutive slices before and after the selected slice were used in the *Selected 3* experiment, and the nine consecutive slices centering on the selected slice were included in the *Selected 9* experiment. The slices obtained through automatic selection (mentioned in Sec. Slice Selection) were all utilized in the *All slices* experiments. These experiments were all conducted with the 2D *Linear* model, and each slice was also copied three times as a three-channel input.

For both data sets, the models performed more favorably when using *Selected 3* and *All slices* than when using *Selected 1* and *Selected 9*. The most favorable performance was observed for the Chang Gung (accuracy: 0.6798; F1 score: 0.5870) and E-Da (accuracy: 0.5248; F1 score: 0.2902) data sets with *All slices* and *Selected 3*, respectively. However, the use of all slices yielded the lowest standard deviations. In terms of the model efficacy and stability, the use of all slices without manual selection had equal or better performance than using partially selected slices. Therefore, the following experiments were performed with all slices after automatic selection.

[Table 2 about here.]

**Training Two Data Sets Separately**

*Chang Gung data set*: The 2D models' experiment results are presented in the top part of Table 3. In general, the models that included age and sex yielded more favorable results (both accuracy and F1 score) than those without them for the Chang Gung data set. Among the 2D models, the *Conv2D* model without age and sex exhibited the highest accuracy (0.6956), whereas the *Linear* model with age and sex yielded the highest F1 score (0.6107).

*E-Da data set*: For the E-Da data set, age and sex were not necessarily useful (resulting in lower average accuracy and F1 score) and somehow increased the instability of the models (as indicated by higher standard deviation). Overall, the *ACS* model outperformed the other 2D models for the E-Da data set, achieving an accuracy of 0.5699 with age and sex and an F1 score of 0.3648 without age and sex.

[Table 3 about here.]

**Cotraining Two Data Sets**

The cotraining technique was applied with our 2D models. We set the cotraining parts in the VGG-16 conv of Fig. 4 as the layers before the third max-pooling layer. The overall experimental results are illustrated in the top part of Table 4. The cotraining did not improve the performance of the Chang Gung data set, except for the *ACS* model. However, for the E-Da data set, most of the models exhibited notable progress following cotraining. The *ACS* model with age and sex yielded the most favorable F1 score (0.3688) up to that point, with the previous highest accuracy being 0.5699, achieved using the separate training model.

[Table 4 about here.]

**3D Models**

**Training Two Data Sets Separately**

*Chang Gung data set*: For the Chang Gung data set, age and sex helped improve the model efficacy (higher accuracy and F1 score) and stability (lower standard deviation) (the middle part of Table 3). The *R(2+1)D* model with age and sex outperformed the other 3D models, with an accuracy of 0.6498 and F1 score of 0.5039. However, the 3D models exhibited poorer performance than the 2D models.

    *E-Da data set*: By contrast, age and sex did not help improve model efficacy for this data set. The *R3D* model without age and sex exhibited the strongest

performance, with an accuracy of 0.4904 and F1 score of 0.3057. However, the 3D models still had poorer accuracy and F1 scores than the 2D models.

**Cotraining Two Data Sets**

The F1 score revealed that the cotraining technique helped improve the model performance, but it did not contribute much in terms of accuracy (the middle part of Table 4). In addition, the inclusion of age and sex was helpful in almost all of our models. For the Chang Gung data set, the *R(2+1)D* model with age and sex achieved an accuracy of 0.6499 and F1 score of 0.5129; for the E-Da data set, the *R3D* model with age and sex achieved an accuracy of 0.5149, and the *R(2+1)D* model with age and sex achieved an F1 score of 0.3595. Despite the improvements resulting from the use of cotraining, the results of the 3D models were still inferior to those of the 2D models.

## Slice-Relation-Based Models

We subsequently considered the relationship among slices in our slice-relation-based models. The experimental results are listed in the bottom part of Tables 3 and 4. In these tables, the suffix "1," as used in "IdxEmb-1" and "Attn-1," indicates that a 2D adaptive average pooling layer with output size $1 \times 1$ is used after the VGG-16 conv; the suffix "4," as used in "IdxEmb-4" and "Attn-4," refers to the absence of an additional 2D pooling layer, yielding an output size for VGG-16 conv of $4 \times 4$.

**Training Two Data Sets Separately**

*Chang Gung data set*: The inclusion of age and sex helped enhance the accuracy and F1 score without increasing the standard deviation for the Chang Gung data set (the bottom part of Table 3). The performance of models with suffix "4" was comparable to that of models with suffix "1." The results indicated that the *Attn-1* model with age and sex yielded the highest accuracy (0.7019) up to that point, and its F1 score (0.5808) was slightly lower than the best F1 score among all models used in this study (0.6107).

*E-Da data set*: For this data set, age and sex were not beneficial to accuracy, but their effect on F1 score was ambiguous because the average and standard deviation both increased. Moreover, the performance of models with suffix "4" was comparable to that of models with suffix "1." The best model here was *IdxEmb-4*, reaching an accuracy of 0.5745 without the addition of age and sex, and an F1 score of 0.3486 with the addition of age and sex.

**Cotraining Two Data Sets**

When cotraining instead of separate model training was employed, model performance for the E-Da data set improved (the bottom part of Table 4). The highest accuracy (0.6766) for the Chang Gung data set was obtained when the *IdxEmb-4* model without age and sex was applied, and the most favorable F1 score (0.5641) was obtained when the *Attn-4* model without age and sex was applied. For the E-Da data set, the highest accuracy (0.5743) was obtained with the *MH-Attn* model without age and sex, and the most favorable F1 score (0.3600) when the *Attn-1* model without age and sex was applied.

**Statistically Significant Performance Difference**

We conducted statistical analyses using one-sided two-sample t-tests with post hoc Bonferroni multiple-comparison correction to evaluate whether the model with the highest metric (accuracy or F1 score) performed significantly better than the others. Our findings revealed that none of the evaluation metrics in Tables 2-4 showed statistically significant differences. As a result, while the bolded values in the tables highlight the models with the highest metric values, the performance differences between models are not statistically significant.

## Discussion

This study presents a comprehensive approach to multiclass classification of PD stages using 3D SPECT brain images. While previous works have utilized 3D CNNs for PD diagnosis [42], our study offers several distinguishing features that contribute to its novelty and practical significance, including:

1. Slice-Relation-Based Models: Previous studies generally treat all slices equally when analyzing 3D volumes or rely on preselected slices. Our proposed slice-relation-based models, which incorporate attention mechanisms, allow the model to assign varying weights to individual slices based on their diagnostic importance. This approach aligns with clinical practices, where physicians focus on slices with clearer striatal patterns.

2. Cotraining Strategy: While prior works have focused on single datasets, we introduce a cotraining strategy to exploit shared characteristics between two independent datasets with different sample sizes and class definitions. This method enhances robustness by increasing sample diversity, a critical factor for generalizing to real-world applications.

3. Demographic Feature Integration: Unlike previous studies, we integrate age and sex into our models, demonstrating that these features significantly improve prediction accuracy, particularly in larger datasets. This highlights the importance of considering patient demographics in PD progression analysis.

4. Thorough Evaluation of 3D Medical Image Classification: Unlike prior works that focus on a single method or architecture, our study systematically explores multiple strategies for leveraging the rich information present in 3D SPECT brain images. This comprehensive evaluation provides valuable insights into the strengths and limitations of different modeling techniques for multiclass classification of PD stages.

Regardless of the model used, the Chang Gung data set consistently outperformed the E-Da data set in terms of accuracy and macro F1 score. This result can be partly attributed to the Chang Gung data set being larger than the E-Da data set (634 vs. 202 samples) and having fewer stages for classification (four vs. six classes). Moreover, compared with the images in the E-Da data set, those in the Chang Gung data set were in a higher resolution (Fig. 3) and the brain occupied a larger proportion of the images. Thus, the superior performance of the Chang Gung data set in the experiments is reasonable.

We used diverse models for different purposes, but many of them did not perform as expected. Theoretically, model performance should increase when more slices are used. Moreover, slices not depicting the striatum should be useless, or even interfere with the predictions. However, the results revealed that using the selected three slices and all slices yielded similar model performance; furthermore, using the selected nine slices yielded less favorable model performance than using the selected three slices. This result may be attributed to the striatum in most of the slices for each patient being insufficiently clear or even absent in the image; by contrast, the selected three slices already contained the most useful information for predictions. Moreover, some of the selected nine slices may interfere with predictions, thus degrading model performance. When all slices without manual selection are used, the information of the whole brain can be retained. Our results suggest that the information gain was greater than the interference caused by not depicting the striatum; thus, no reduction in model performance was noted.

In most cases, the 2D models outperformed the 3D models. This result may be attributable to the different data sets used for the pretrained model: ImageNet was used for the 2D models and Kinetics-400 for the 3D models. ImageNet contains more than 1 million images, whereas Kinetics-400 contains only 306,000 video clips. The use of a large data set during pretraining contributed to more favorable or robust model performance. Moreover, our data were essentially a sequence of static images. Kinetics-400 contains action video clips, and thus, ImageNet was more similar to our target data sets.

In theory, the importance or contribution of each slice in one patient should not be equal; thus, the slice-relation-based models that allowed for different contributions from each slice should outperform the models that do not consider these relations. The results suggested that these slice-relation-based models were appropriate for both data sets because the average accuracy and F1 scores were higher than or equivalent to those for the models not considering these relations. However, the stability decreased (i.e., standard deviation increased) when slice-relation-based models were used. The growth of model complexity may be responsible for this phenomenon because the trainable weights increased in these models. When we used the models without applying cotraining in the experiments, the strongest performance for the Chang Gung and E-Da data sets was obtained with the *Attn-1* and *IdxEmb-4* models, respectively. Therefore, the relations among slices are crucial and should be considered when making predictions.

We anticipated that model efficacy and robustness for both data sets would improve after applying cotraining because of the increase in sample size. However, this technique did not consistently help improve model performance: the improvement was evident for the E-Da data set but ambiguous for the Chang Gung data set. The sample size of the E-Da data set was insufficient to support our deep learning models by itself. The inclusion of the Chang Gung data set during training helped improve the model performance in the target E-Da data set because of the increase in sample size and diversity. The Chang Gung data set had more samples and higher resolution images for training and fewer stages to be classified; thus, cotraining the smaller E-Da data set for the more complex classification task may adversely affect the model performance in the Chang Gung data set. In summary, the cotraining technique was useful when the data set used for cotraining had a sufficiently large sample size and its corresponding task was sufficiently similar to the target task.

When training two data sets separately, age and sex generally helped improve the accuracy and F1 score in the Chang Gung data set but not in the E-Da data set. When we cotrained the two datasets, including age and sex in the model became beneficial for both data sets. The adverse effect on model fitting caused by the small sample size of the E-Da data set seemed to escalate when fitting a more complex model that included age and sex. Thus, we conclude that age and sex can help the stage prediction of PD as long as the training data are of sufficient size to support the deep learning models used.

While our results are encouraging, certain limitations must be acknowledged. For example, the cotraining strategy produced mixed results for the Chang Gung dataset, highlighting the need for further refinement to optimize its performance. Additionally, the E-Da dataset suffers from significant class imbalance, which can lead to biased models that favor majority classes. To address this, future work could explore synthetic data generation methods, such as Generative Adversarial Networks (GANs) [43] and class-specific sampling techniques like the Synthetic Minority Over-sampling Technique (SMOTE) [44]. These strategies could enhance the representation of under-represented classes and improve overall model balance. Moreover, the relatively small overall dataset size, particularly for the E-Da cohort, may limit the robustness and generalizability of the models. Integrating larger, publicly available datasets, such as the PPMI cohort [45], could provide more diverse data, thereby increasing the robustness of the proposed models and more effectively addressing class imbalance issues.

## Conclusion

In this study, we designed and developed various model architectures for predicting the stages of PD, a multiclass classification task. Moreover, we used the whole 3D brain image from each patient for prediction instead of selecting one or some of the slices; this approach enabled us to save time on manual selection, and the overall procedure or pipeline was fully automatic. We then applied trilinear interpolation to address the different number of slices for each patient, and we included class weights in the loss function because of the imbalanced data. Among the models we used, the 2D models pretrained on ImageNet outperformed the 3D models pretrained on Kinetics-400. Moreover, considering the relations among slices was beneficial for model performance and did not excessively increase the number of trainable weights. The use of the cotraining technique is one of the key contributions in our study. Cotraining helped improve the model efficacy and robustness under some constraints such as small sample size and dissimilarity between the data sets. Moreover, age and sex aided the stage prediction of PD, but only when the training data were sufficiently large. In conclusion, our study significantly advances the field by combining innovative modeling techniques, leveraging demographic data, and introducing a robust pipeline for multiclass PD stage classification. These contributions provide a strong foundation for future research and clinical translation.

# References

1. Tysnes OB, Storstein A: Epidemiology of Parkinson's disease. J Neural Transm 124(8):901-905, 2017

2. Liu WM, Wu RM, Lin JW, Liu YC, Chang CH, Lin CH: Time trends in the prevalence and incidence of Parkinson's disease in Taiwan: A nationwide, population-based study. J Formos Med Assoc 115(7):531-538, 2016

3. Hoehn MM, Yahr MD: Parkinsonism: onset, progression and mortality. Neurology 17(5):427-442, 1967

4. Parkinson's Foundation. Available at https://www.parkinson.org/Understanding-Parkinsons/What-is-Parkinsons/Stages-of-Parkinsons. Accessed 3 May 2023.

5. Felicio AC, Shih MC, Godeiro-Junior C, Andrade LAF, Bressan RA, Ferraz HB: Molecular imaging studies in Parkinson disease: reducing diagnostic uncertainty. Neurologist 15(1):6-16, 2009

6. Scherfler C, Nocker M: Dopamine transporter SPECT: how to remove subjectivity? Mov Disord 24(S2):S721-S724, 2009

7. Staff RT, Ahearn TS, Wilson K, Counsell CE, Taylor K, Caslake R, Davidson JE, Gemmell HG, Murray AD: Shape analysis of 123I-N-omega-fluoropropyl-2-beta-carbomethoxy-3beta-(4-iodophenyl) nortropane single-photon emission computed tomography images in the assessment of patients with parkinsonian

syndromes. Nucl Med Commun 30(3):194-201, 2009

8. Prashanth R, Roy SD, Mandal PK, Ghosh S: High-accuracy classification of Parkinson's disease through shape analysis and surface fitting in 123I-Ioflupane SPECT imaging. IEEE J Biomed Health Inform 21(3):794-802, 2016

9. Towey DJ, Bain PG, Nijran KS: Automatic classification of 123I-FP-CIT (DaTSCAN) SPECT images. Nucl Med Commun 32(8):699-707, 2011

10. Illán I, Górriz J, Ramírez J, Segovia F, Jiménez-Hoyuela J, Ortega Lozano S: Automatic assistance to Parkinson's disease diagnosis in DaTSCAN SPECT imaging. Med Phys 39(10):5971-5980, 2012

11. Segovia F, Górriz JM, Ramírez J, Alvarez I, Jiménez-Hoyuela JM, Ortega SJ: Improved parkinsonism diagnosis using a partial least squares based approach. Med Phys 39(7):4395-4403, 2012

12. Rojas A, Górriz J, Ramírez J, Illán I, Martínez-Murcia FJ, Ortiz A, Río MG, Moreno-Caballero M: Application of Empirical Mode Decomposition (EMD) on DaTSCAN SPECT images to explore Parkinson disease. Expert Syst Appl 40(7):2756-2766, 2013

13. Prashanth R, Roy SD, Mandal PK, Ghosh S: Automatic classification and prediction models for early Parkinson's disease diagnosis from SPECT imaging. Expert Syst Appl 41(7):3333-3342, 2014

14. Bhalchandra NA, Prashanth R, Roy SD, Noronha S: Early detection of Parkinson's disease through shape based features from 123I-Ioflupane SPECT imaging. In: IEEE 12th International Symposium on Biomedical Imaging (ISBI), pp. 963-966, 2015

15. Pagan FL: Improving outcomes through early diagnosis of Parkinson's disease. Am J Manag Care 18(7):S176, 2012

16. Caesarendra W, Ariyanto M, Setiawan JD, Arozi M, Chang CR: A pattern recognition method for stage classification of Parkinson's disease utilizing voice features. In: IEEE Conference on Biomedical Engineering and Sciences (IECBES), pp. 87-92, 2014

17. Choi H, Ha S, Im HJ, Paek SH, Lee DS: Refining diagnosis of Parkinson's disease with deep learning based interpretation of dopamine transporter imaging. Neuroimage Clin 16:586-594, 2017

18. Magesh PR, Myloth RD, Tom RJ: An explainable machine learning model for early detection of Parkinson's disease using LIME on DaTSCAN imagery. Comput Biol Med 126:104041, 2020

19. Chien C-Y, Hsu S-W, Lee T-L, Sung P-S, Lin C-C: Using artificial neural network to discriminate Parkinson's disease from other Parkinsonisms by focusing on putamen of dopamine transporter SPECT images. Biomedicines 9(1):12, 2020

20. Huang GH, Lin CH, Cai YR, Chen TB, Hsu SY, Lu NH, Chen HY, Wu YC: Multiclass machine learning classification of functional brain images for Parkinson's disease stage prediction. Stat Anal Data Min 13(5):508-523, 2020

21. Karpathy A, Toderici G, Shetty S, Leung T, Sukthankar R, Fei-Fei L: Large-scale video classification with convolutional neural networks. In: IEEE Conference on Computer Vision and Pattern Recognition, pp. 1725-1732, 2014

22. Setio AAA, Ciompi F, Litjens G, Gerke P, Jacobs C, van Riel SJ, Wille MMW, Naqibullah M, Sanchez CI, van Ginneken B: Pulmonary nodule detection in ct images: false positive reduction using multi-view convolutional networks. IEEE Trans Med Imaging 35(5):1160-1169, 2016

23. Tran D, Wang H, Torresani L, Ray J, LeCun Y, Paluri M: A closer look at

spatiotemporal convolutions for action recognition. In: Proceedings of the IEEE conference on Computer Vision and Pattern Recognition, pp. 6450-6459, 2018

24. Singh SP, Wang L, Gupta S, Goli H, Padmanabhan P, Gulyás B: 3D deep learning on medical images: a review. *Sensors* 20:5097, 2020

25. Yang J, Huang X, He Y, Xu J, Yang C, Xu G, Ni B: Reinventing 2D convolutions for 3D images. arXiv preprint arXiv:1911.10477, 2019

26. Huang SF, Wen YH, Chu CH, Hsu CC: A shape approximation for medical imaging data. Sensors 20:5879, 2020

27. Wooten GF, Currie LJ, Bovbjerg VE, Lee JK, Patrie J: Are men at greater risk for Parkinson's disease than women? J Neurol Neurosurg Psychiatry 75(4):637-639, 2004

28. Pringsheim T, Jette N, Frolkis A, Steeves TD: The prevalence of Parkinson's disease: a systematic review and meta-analysis. Mov Disord 29(13):1583-1590, 2014

29. Rajon DA, Bolch WE: Marching cube algorithm: review and trilinear interpolation adaptation for image-based dosimetric models. Comput Med Imaging Graph 27(5):411-435, 2003

30. Krizhevsky A, Sutskever I, Hinton GE: ImageNet classification with deep convolutional neural networks. Commun ACM 60:84-90, 2017

31. Kay W, Carreira J, Simonyan K, Zhang B, Hillier C, Vijayanarasimhan S, Viola F, Green T, Back T, Natsev P, Suleyman M, Zisserman M: The kinetics human action video dataset. arXiv preprint arXiv:1705.06950, 2017

32. Tajbakhsh N, Shin JY, Gurudu SR, Hurst RT, Kendall CB, Gotway MB, Liang J: Convolutional neural networks for medical image analysis: Full training or fine tuning? IEEE Trans Med Imaging 35:1299-1312, 2016

33. Huang GH, Fu QJ, Gu MZ, Lu NH, Liu KY, Chen TB: Deep transfer learning for the multilabel classification of chest X-ray images. Diagnostics 12:1457, 2020

34. Liu S, Xu D, Zhou SK, Pauly O, Grbic S, Mertelmeier T, Wicklein J, Jerebko A, Cai W, Comaniciu D: 3d anisotropic hybrid network: Transferring convolutional features from 2d images to 3d anisotropic volumes. In: International Conference on Medical Image Computing and Computer-Assisted Intervention, pp. 851-858, 2018

35. Guo C, Berkhahn F: Entity embeddings of categorical variables. arXiv preprint arXiv:1604.06737, 2016

36. Bahdanau D, Cho K, Bengio Y: Neural machine translation by jointly learning to align and translate. arXiv preprint arXiv:1409.0473, 2014

37. Luong M-T, Pham H, Manning CD: Effective approaches to attention-based neural machine translation. arXiv preprint arXiv:1508.04025, 2015

38. Lin Z, Feng M, Nogueira dos Santos C, Yu M, Xiang B, Zhou B, Bengio Y: A structured self-attentive sentence embedding. arXiv preprint arXiv:1703.03130, 2017

39. Vaswani A, Shazeer N, Parmar N, Uszkoreit J, Jones L, Gomez AN, Kaiser L, Polosukhin I: Attention is all you need. arXiv preprint arXiv:1706.03762, 2017

40. Smith LN, Topin N: Super-convergence: very fast training of neural networks using large learning rates. arXiv preprint arXiv:1708.07120, 2017

41. Loshchilov I, Hutter F: SGDR: Stochastic gradient descent with warm restarts. In: 5th International Conference on Learning Representations (ICLR), 2017

42. Tufail AB, Ma YK, Zhang QN, Khan A, Zhao L, Yang Q, Adeel M, Khan R, Ullah I: 3D convolutional neural networks-based multiclass classification of Alzheimer's and Parkinson's diseases using PET and SPECT neuroimaging modalities. Brain Inf 8:23, 2021

43. Huang G, Jafari AH: Enhanced balancing GAN: minority-class image generation. Neural Comput Appl 5:5145-5154, 2023

44. Hasan Y, Amerehi F, Healy P, Ryan C: STEM rebalance: A novel approach for tackling imbalanced datasets using SMOTE, edited nearest neighbour, and mixup. arXiv preprint arXiv:2311.07504, 2023

45. Parkinson Progression Marker Initiative: The Parkinson Progression Marker Initiative (PPMI). Prog Neurobiol 95(4):629-635, 2011

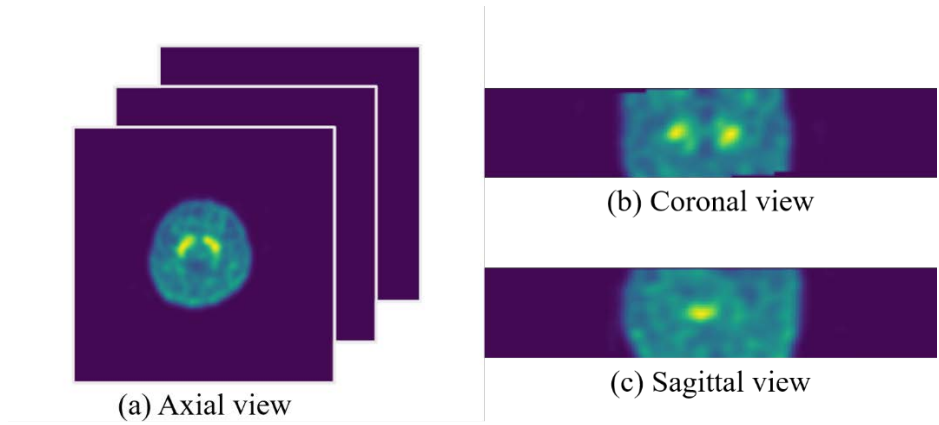

**Fig. 1**　Three views of our 3D SPECT imaging: (a) axial view, (b) coronal view, and (c) sagittal view

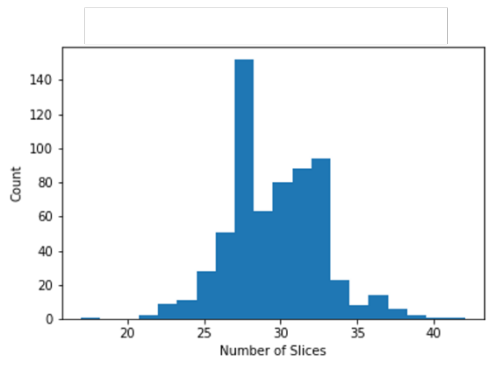 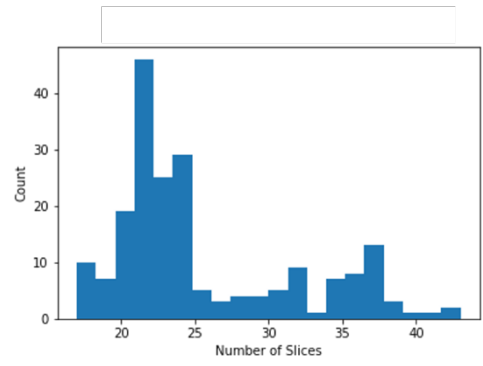

(a) Chang Gung dataset  (b) E-Da dataset

**Fig. 2**  Bar plots of the number of slices for our 3D SPECT imaging

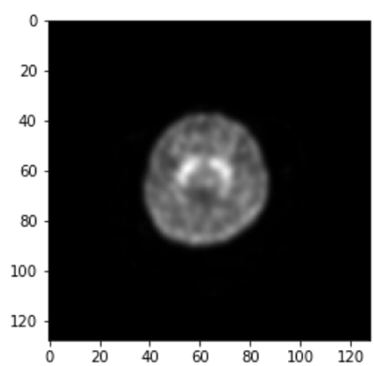 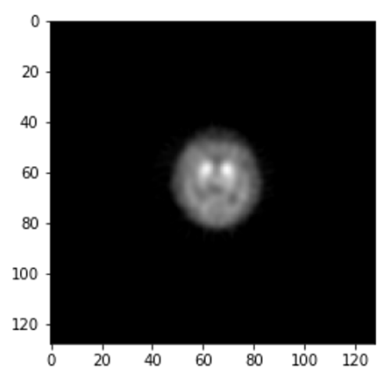

(a) Sample in Chang Gung    (b) Sample in E-Da

**Fig. 3**   Image samples from two target data sets

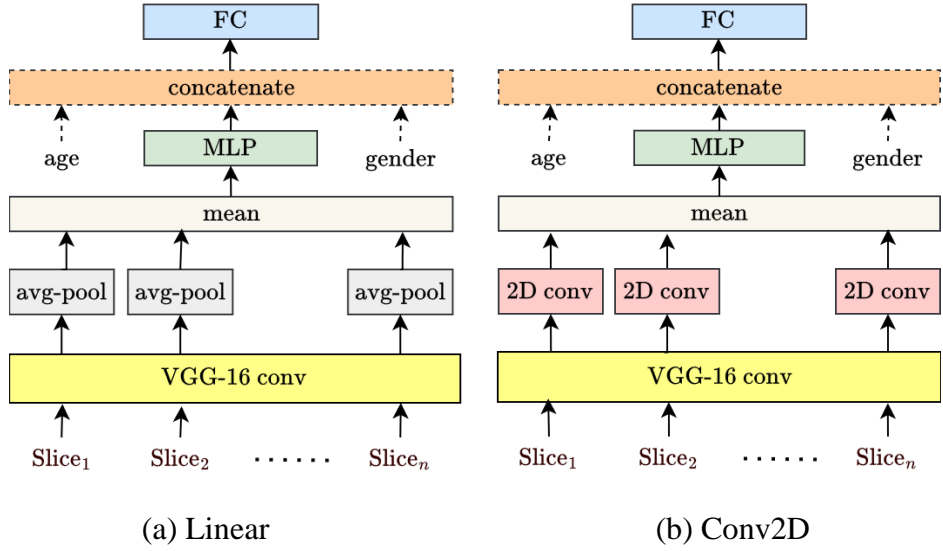
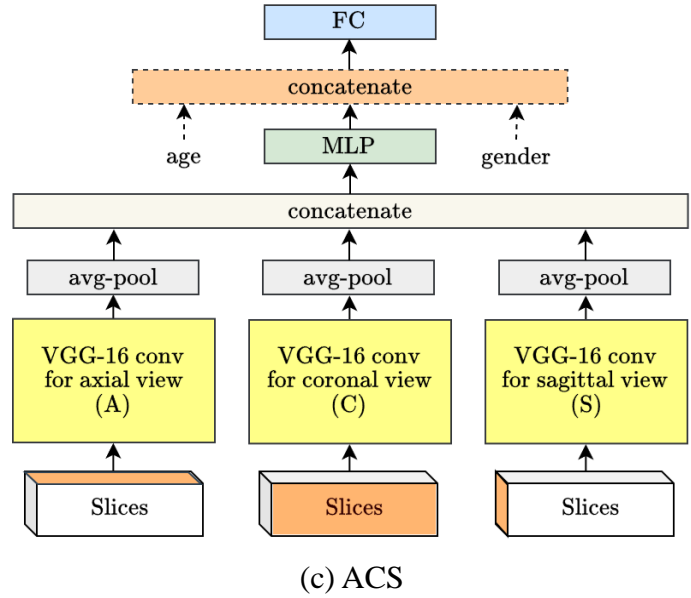

**Fig. 4** 2D model architecture. The yellow block represents the modified VGG-16 architecture we used. In (c), "Slices" represents the collection of all slice sets form the same person

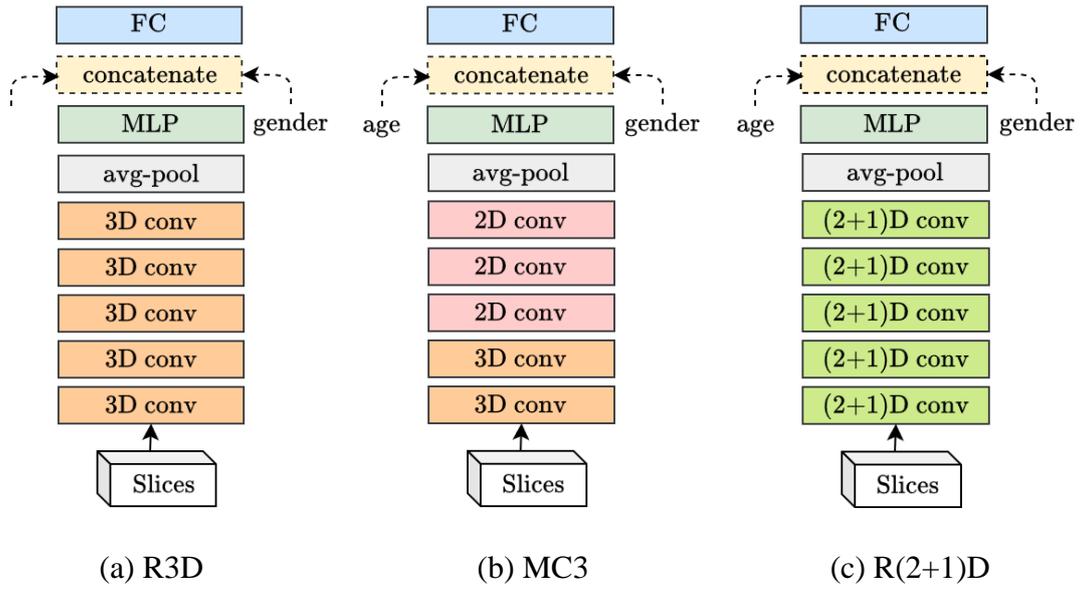

(a) R3D        (b) MC3        (c) R(2+1)D

**Fig. 5**    3D model architecture

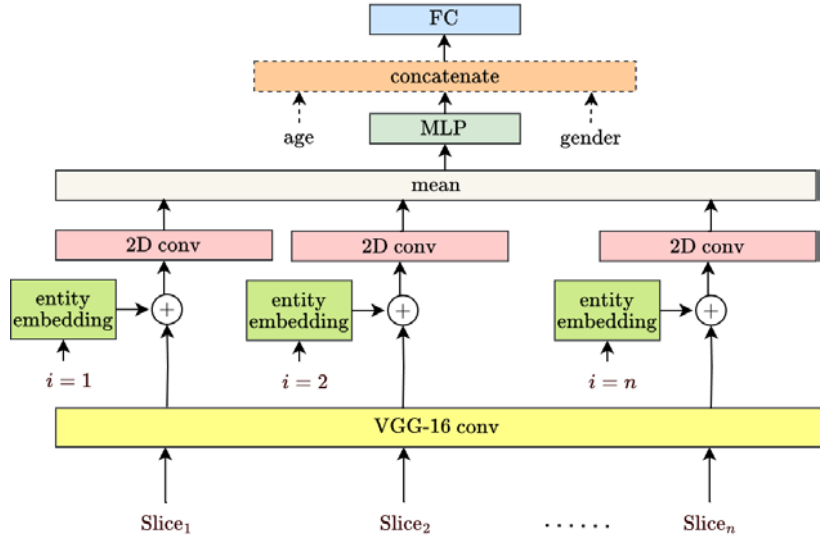

(a) IdxEmb

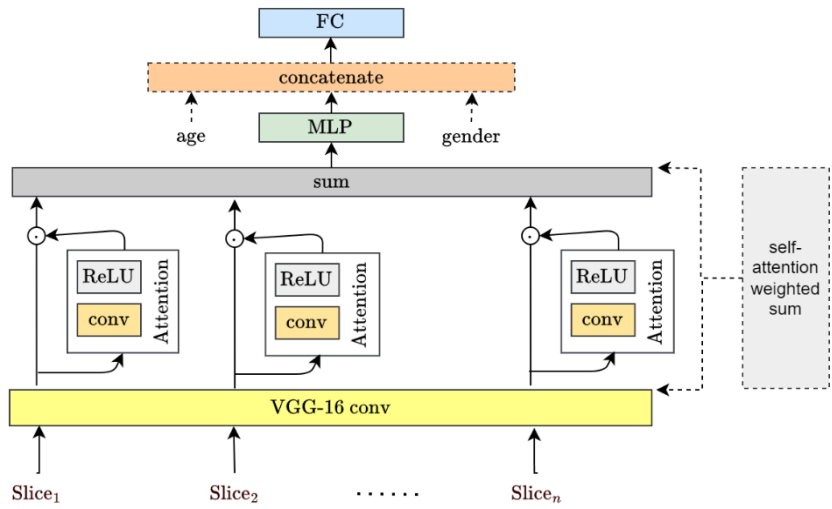

(b) Attn

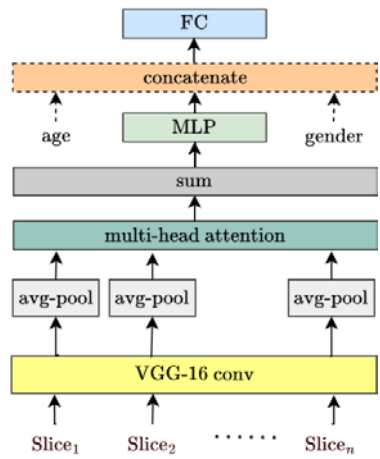

(c) MH-Attn

**Fig. 6**　　Slice-relation-based model architecture

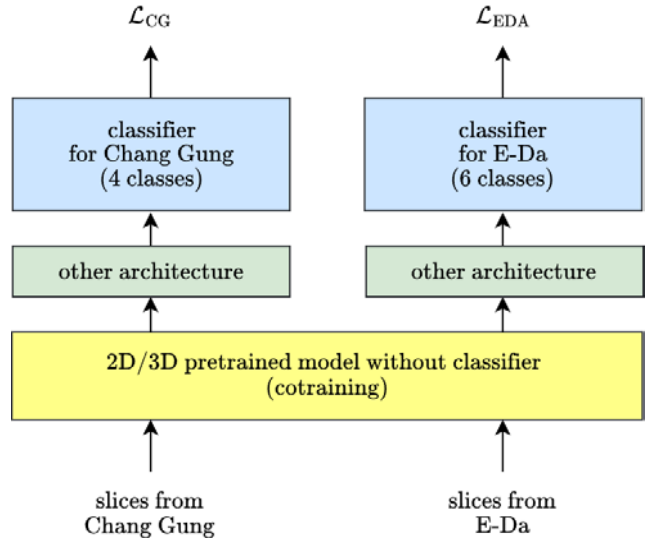

**Fig. 7**    Schematic of cotraining models

**Table 1** Summary of experimental configurations and parameter settings

| Configuration | Setting |
|---|---|
| Preprocessing | |
|     Normalization | Min-max normalization for each image |
|     Selection of SPECT imaging slices for analysis | Manual selection by radiology specialists or removal of slices whose number of pixels with normalized intensity value greater than 0.1 was less than 800 and 400 for the Chang Gung and E-Da data sets, respectively |
| Augmentation | Online augmentation to generate images of size (32,72,72) by using video transform and trilinear interpolation for depth |
| Loss function | Weighted categorical cross-entropy |
| Deep CNN modeling | |
|     2D model | Linear, Conv2D, and ACS |
|     3D model | R3D, MC3, and R(2+1)D |
|     Slice-relation-based model | IdxEmb, Attn, and MH-Attn |
|     Optimizer | Adam |
|     Scheduler | One cycle policy [40] |
|     Mini-batch size | 8 |
|     Number of steps | 3000 |
|     Learning rate | Start from 0.0001 and drop with the cosine annealing strategy [41] |
| Transfer learning | |
|     2D model architecture | Pre-train on ImageNet |
|     3D model architecture | Pre-train on Kinetics400 |
| Evaluation | |
|     Data splitting | Stratified 5-fold cross-validation |
|     Metrics | Accuracy, macro F1 score |

**Table 2** Experimental results for different number of slices used

| # slices | Chang Gung | | E-Da | |
|---|---|---|---|---|
| | Accuracy[a] | F1 score[b] | Accuracy[a] | F1 score[b] |
| Selected 1 | 0.6356 (±0.03) | 0.5118 (±0.08) | 0.4898 (±0.07) | 0.2503 (±0.05) |
| Selected 3 | 0.6750 (±0.04) | 0.5675 (±0.06) | **0.5248** (±0.06) | **0.2902** (±0.07) |
| Selected 9 | 0.6687 (±0.03) | 0.5294 (±0.05) | 0.4305 (±0.06) | 0.2052 (±0.05) |
| All slices | **0.6798** (±0.03) | **0.5870** (±0.04) | 0.5000 (±0.04) | 0.2445 (±0.04) |

[a] Average (±standard deviation) test accuracy of five test folds

[b] Average (±standard deviation) test macro F1 score of five test folds

Values in boldface denote the best result (the highest value) in each evaluation metric

**Table 3** Experimental results for separately trained models

| | | **Chang Gung** | | **E-Da** | |
|---|---|---|---|---|---|
| **Model**[a] | | Accuracy[b] | F1 score[c] | Accuracy[b] | F1 score[c] |
| Linear | − | 0.6798 (±0.03) | 0.5870 (±0.03) | 0.5000 (±0.07) | 0.2445 (±0.04) |
| | + | 0.6940 (±0.03) | **0.6107** (±0.03) | 0.4751 (±0.09) | 0.2974 (±0.11) |
| Conv2D | − | 0.6956 (±0.02) | 0.5567 (±0.06) | 0.4700 (±0.04) | 0.2363 (±0.04) |
| | + | 0.6862 (±0.04) | 0.5570 (±0.06) | 0.4606 (±0.03) | 0.1987 (±0.02) |
| ACS | − | 0.6057 (±0.02) | 0.4963 (±0.04) | 0.5294 (±0.04) | **0.3648** (±0.09) |
| | + | 0.6198 (±0.02) | 0.5105 (±0.04) | 0.5699 (±0.06) | 0.3396 (±0.07) |
| R3D | − | 0.6293 (±0.03) | 0.4626 (±0.03) | 0.4904 (±0.07) | 0.3057 (±0.07) |
| | + | 0.6388 (±0.02) | 0.4757 (±0.04) | 0.4656 (±0.06) | 0.2600 (±0.09) |
| MC3 | − | 0.6372 (±0.02) | 0.4664 (±0.05) | 0.4851 (±0.04) | 0.2453 (±0.04) |
| | + | 0.6372 (±0.03) | 0.4914 (±0.03) | 0.4409 (±0.03) | 0.2526 (±0.04) |
| R(2+1)D | − | 0.6467 (±0.04) | 0.4703 (±0.05) | 0.4800 (±0.04) | 0.2849 (±0.06) |
| | + | 0.6498 (±0.01) | 0.5039 (±0.03) | 0.4610 (±0.06) | 0.2471 (±0.08) |
| IdxEmb-1 | − | 0.6529 (±0.03) | 0.5017 (±0.08) | 0.5098 (±0.06) | 0.3074 (±0.10) |
| | + | 0.6671 (±0.03) | 0.5605 (±0.05) | 0.4760 (±0.11) | 0.2792 (±0.08) |
| IdxEmb-4 | − | 0.6703 (±0.04) | 0.5800 (±0.06) | **0.5745** (±0.06) | 0.3022 (±0.03) |
| | + | 0.6750 (±0.03) | 0.5467 (±0.08) | 0.5300 (±0.04) | 0.3486 (±0.04) |
| Attn-1 | − | 0.6814 (±0.02) | 0.5390 (±0.04) | 0.5445 (±0.06) | 0.2859 (±0.06) |
| | + | **0.7019** (±0.02) | 0.5808 (±0.04) | 0.5100 (±0.03) | 0.3117 (±0.06) |
| Attn-4 | − | 0.6781 (±0.04) | 0.5458 (±0.07) | 0.5395 (±0.07) | 0.3006 (±0.08) |
| | + | 0.6703 (±0.01) | 0.5674 (±0.01) | 0.5448 (±0.06) | 0.3137 (±0.09) |
| MH-Attn | − | 0.6481 (±0.05) | 0.5021 (±0.05) | 0.5249 (±0.06) | 0.3079 (±0.12) |
| | + | 0.6641 (±0.02) | 0.5179 (±0.03) | 0.5246 (±0.07) | 0.3385 (±0.08) |

[a] −: age and sex are not included as inputs; +: age and sex are concatenated as inputs. The suffix "1," as used in "IdxEmb-1" and "Attn-1," indicates that a 2D adaptive average pooling layer with output size $1 \times 1$ is used after the VGG-16 conv; the suffix "4," as used in "IdxEmb-4" and "Attn-4," refers to the absence of an additional 2D pooling layer, yielding an output size for VGG-16 conv of $4 \times 4$.
[b] average (±standard deviation) test accuracy of five test folds
[c] average (±standard deviation) test macro F1 score of five test folds
Values in boldface denote the best result (the highest value) in each evaluation metric

**Table 4** Experimental results for cotrained models

| Model[a] | | Chang Gung | | E-Da | |
|---|---|---|---|---|---|
| | | Accuracy[b] | F1 score[c] | Accuracy[b] | F1 score[c] |
| Linear | − | 0.6703 (±0.03) | 0.5535 (±0.04) | *0.5299 (±0.03) | *0.2953 (±0.03) |
| | + | 0.6529 (±0.02) | 0.5032 (±0.05) | *0.5599 (±0.05) | *0.3191 (±0.09) |
| Conv2D | − | 0.6750 (±0.03) | 0.5430 (±0.06) | *0.5450 (±0.06) | *0.3306 (±0.08) |
| | + | 0.6655 (±0.04) | 0.5385 (±0.08) | *0.5640 (±0.05) | *0.2910 (±0.05) |
| ACS | − | *0.6403 (±0.03) | *0.5252 (±0.06) | *0.5695 (±0.04) | 0.3181 (±0.04) |
| | + | *0.6244 (±0.06) | *0.5129 (±0.08) | 0.5495 (±0.05) | ***0.3688** (±0.05) |
| R3D | − | *0.6294 (±0.02) | *0.4712 (±0.04) | 0.4704 (±0.04) | 0.2743 (±0.07) |
| | + | *0.6498 (±0.02) | *0.5046 (±0.03) | *0.5149 (±0.05) | *0.3039 (±0.04) |
| MC3 | − | 0.6214 (±0.03) | *0.4808 (±0.03) | *0.5146 (±0.04) | *0.2979 (±0.08) |
| | + | 0.6340 (±0.04) | *0.4880 (±0.02) | *0.5100 (±0.03) | *0.3093 (±0.11) |
| R(2+1)D | − | 0.6419 (±0.02) | *0.4777 (±0.04) | 0.4507 (±0.05) | *0.2941 (±0.09) |
| | + | *0.6499 (±0.03) | *0.5129 (±0.05) | *0.5093 (±0.08) | *0.3595 (±0.08) |
| IdxEmb-1 | − | 0.6103 (±0.03) | 0.4854 (±0.03) | *0.5589 (±0.04) | *0.3583 (±0.10) |
| | + | 0.6466 (±0.04) | 0.5216 (±0.05) | *0.5440 (±0.08) | *0.3351 (±0.13) |
| IdxEmb-4 | − | ***0.6766** (±0.04) | 0.5578 (±0.06) | 0.5546 (±0.02) | *0.3109 (±0.08) |
| | + | 0.6513 (±0.03) | 0.5423 (±0.07) | 0.5052 (±0.05) | 0.2905 (±0.04) |
| Attn-1 | − | 0.6308 (±0.04) | 0.5086 (±0.06) | *0.5595 (±0.03) | *0.3600 (±0.10) |
| | + | 0.6750 (±0.04) | 0.5489 (±0.07) | *0.5394 (±0.05) | 0.3026 (±0.02) |
| Attn-4 | − | 0.6639 (±0.05) | ***0.5641** (±0.06) | *0.5449 (±0.05) | *0.3085 (±0.05) |
| | + | 0.6608 (±0.04) | 0.5331 (±0.08) | 0.5101 (±0.09) | *0.3158 (±0.10) |
| MH-Attn | − | 0.6309 (±0.02) | 0.4837 (±0.03) | ***0.5743** (±0.02) | *0.3505 (±0.05) |
| | + | 0.6639 (±0.04) | *0.5455 (±0.05) | *0.5594 (±0.05) | *0.3471 (±0.11) |

[a] −: age and sex are not included as inputs; +: age and sex are concatenated as inputs. The suffix "1," as used in "IdxEmb-1" and "Attn-1," indicates that a 2D adaptive average pooling layer with output size $1 \times 1$ is used after the VGG-16 conv; the suffix "4," as used in "IdxEmb-4" and "Attn-4," refers to the absence of an additional 2D pooling layer, yielding an output size for VGG-16 conv of $4 \times 4$.

[b] average (±standard deviation) test accuracy of five test folds

[c] average (±standard deviation) test macro F1 score of five test folds

* better performance than when trained separately

Values in boldface denote the best result (the highest value) in each evaluation metric